\documentclass[conference]{IEEEtran}
\usepackage{amsmath,amsfonts}
\usepackage{algorithmic}
\usepackage{algorithm}
\usepackage{array}
\usepackage{subcaption}

\usepackage{textcomp} 
\usepackage{stfloats}
\usepackage[hyphens]{url}
\usepackage{verbatim}
\usepackage{graphicx}
\usepackage[dvipsnames]{xcolor}
\usepackage[numbers]{natbib}
\usepackage{times}
\usepackage{multicol}
\PassOptionsToPackage{hyphens}{url}\usepackage{hyperref}

\let\labelindent\relax
\usepackage{enumitem}
\usepackage{tabularx}

\usepackage{booktabs}       

\usepackage{arydshln}
\makeatletter
\def\adl@drawiv#1#2#3{%
        \hskip.5\tabcolsep
        \xleaders#3{#2.5\@tempdimb #1{1}#2.5\@tempdimb}%
                #2\z@ plus1fil minus1fil\relax
        \hskip.5\tabcolsep}
\newcommand{\cdashlinelr}[1]{%
  \noalign{\vskip\aboverulesep
           \global\let\@dashdrawstore\adl@draw
           \global\let\adl@draw\adl@drawiv}
  \cdashline{#1}
  \noalign{\global\let\adl@draw\@dashdrawstore
           \vskip\belowrulesep}}
\makeatother

\newcommand{\textmidtilde}{\raisebox{0.5ex}{\texttildelow}}

\def\publishedurl{\centering \textbf{{\color{NavyBlue}{Robotics: Science and Systems (RSS 2023)}\color{ProcessBlue}{, July 10 - 14, 2023, Daegu, Republic of Korea} \\\color{red}{The published version can be accessed from}} \color{blue}{\url{https://roboticsconference.org/program/papers/023}}}}

\def\papertitle{Demonstrating Large-Scale Package Manipulation\\ via Learned Metrics of Pick Success}

\begin{document}

\title{\papertitle}

\author{\authorblockN{Shuai Li\authorrefmark{1}, Azarakhsh Keipour\authorrefmark{1}, Kevin Jamieson\authorrefmark{1}\authorrefmark{2}, Nicolas Hudson\authorrefmark{1}, Charles Swan\authorrefmark{1} and Kostas Bekris\authorrefmark{1}\authorrefmark{3}}
\authorblockA{\authorrefmark{1}Amazon Robotics, Seattle, Washington 98109, USA\\
\authorrefmark{2}University of Washington, Seattle, Washington 98105, USA\\
\authorrefmark{3}Rutgers University, Piscataway, New Jersey 08854, USA \\
Email: \{amzshua, keipourv, jamikevi, hudnco, cswan, bekris\}@amazon.com}
}

\maketitle

\begin{abstract}
Automating warehouse operations can reduce logistics overhead costs, ultimately driving down the final price for consumers, increasing the speed of delivery, and enhancing the resiliency to workforce fluctuations. The past few years have seen increased interest in automating such repeated tasks but mostly in controlled settings. Tasks such as picking objects from unstructured, cluttered piles have only recently become robust enough for large-scale deployment with minimal human intervention.

This paper demonstrates a large-scale package manipulation from unstructured piles in Amazon Robotics' Robot Induction (Robin) fleet, which utilizes a pick success predictor trained on real production data. Specifically, the system was trained on over 394K picks. It is used for singulating up to 5~million packages per day and has manipulated over 200~million packages during this paper's evaluation period.

The developed learned pick quality measure ranks various pick alternatives in real-time and prioritizes the most promising ones for execution. The pick success predictor aims to estimate from prior experience the success probability of a desired pick by the deployed industrial robotic arms in cluttered scenes containing deformable and rigid objects with partially known properties. It is a shallow machine learning model, which allows us to evaluate which features are most important for the prediction. An online pick ranker leverages the learned success predictor to prioritize the most promising picks for the robotic arm, which are then assessed for collision avoidance. This learned ranking process is demonstrated to overcome the limitations and outperform the performance of manually engineered and heuristic alternatives.

To the best of the authors' knowledge, this paper presents the first large-scale deployment of learned pick quality estimation methods in a real production system.

\end{abstract}

\IEEEpeerreviewmaketitle

\section{Introduction} \label{sec:intro}

Automation in the industrial, manufacturing, and warehouse sectors has the potential to lower overhead expenses associated with producing, handling, and sorting goods. The increased speed and precision for handling each product can lower customer costs and improve product quality. Furthermore, it can reduce risks to humans in manual operations and enhance resilience to fluctuations in the labor market and overall economy. 

\begin{figure}[t]
    \centering
    \includegraphics[width=0.497\linewidth, height=5cm]{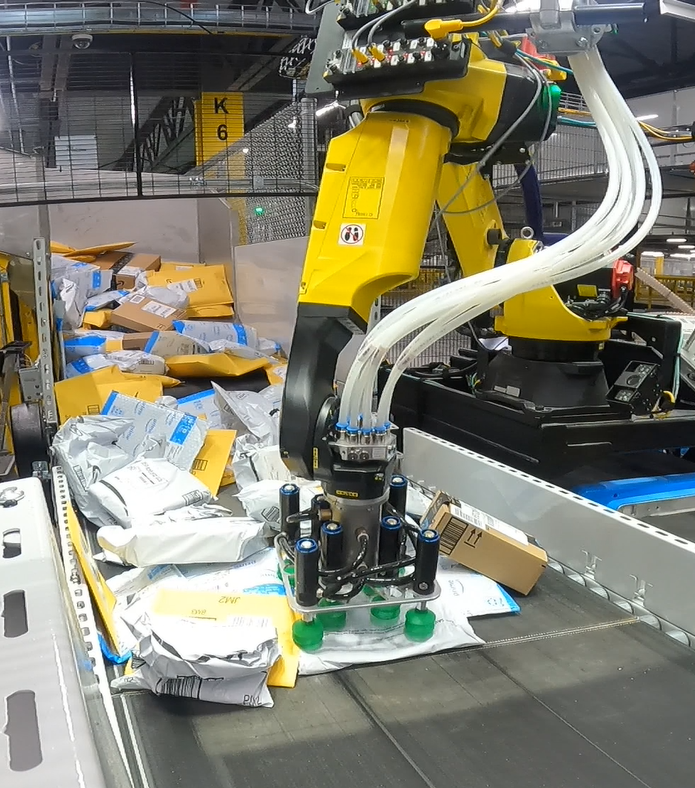}%
    \hfill%
    \includegraphics[width=0.497\linewidth, height=5cm]{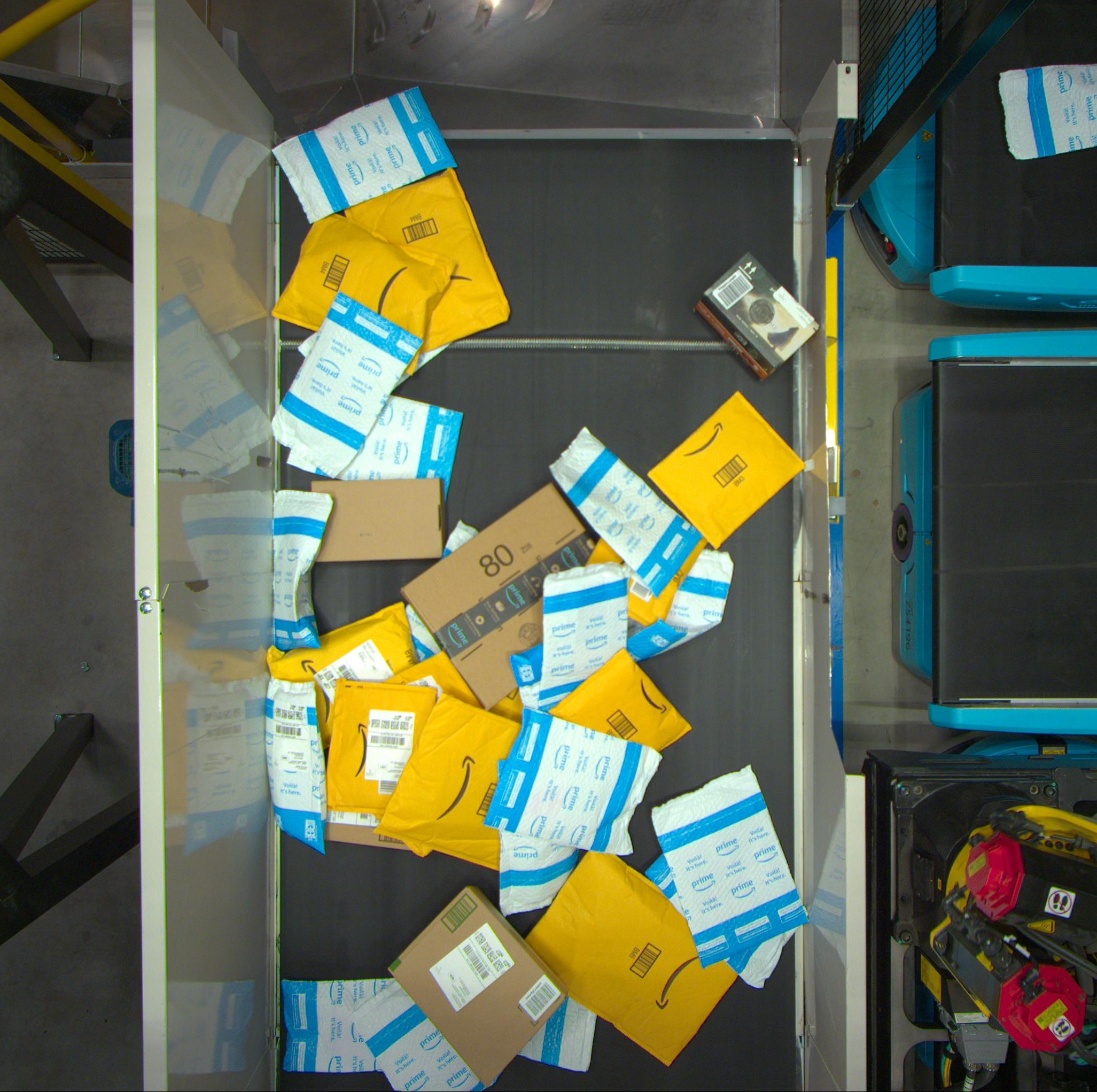}%
    \caption{A robot induction (Robin) workcell used for the statistics of this demonstration. The robotic arm is used for automated package singulation by Amazon.com, Inc. It picks packages from an unstructured pile on a conveyor belt and places them on mobile drive robots.}
    \vspace{-.2in}
    \label{fig:robin-workcell}
\end{figure}

Robot manipulation systems have already gained significant traction across industries, from car and garment manufacturing to crating apples~\cite{garment-automation, car-wiring, apple-binning-review, apple-crating}. Many repeated operations in industrial settings include pick-and-place tasks using robot arms~\cite{pick-and-place-automation, pick-and-place-manufacturing}. Induction robots pick items from one location (e.g., a conveyor belt, a tote, or a box) and place them in another spot with the goal of singulating the items or feeding them to another machine (e.g., a sortation machine). Automation is still far from perfect, however. Some robot settings require a simplified environment to operate (e.g., only a single rigid object placed around the center of the conveyor belt), while others can only deal with a subset of the target objects, and the rest are passed on to humans (e.g., apple crating). Success metrics and cost benefits vary across tasks, and while these systems are largely beneficial, recent advances in robotics, computer vision, and machine learning are providing additional opportunities for robots to become financially viable in rather complex manipulation operations. 

This work presents the learned pick quality system used in the Robot Induction (Robin) fleet of Amazon.com, Inc., which sorts several million packages per day~\cite{robin}. Figure~\ref{fig:robin-workcell} shows a Robin workcell picking packages from a conveyor belt, which has been used for the statistics presented in this demonstration. Once a package is picked, Robin scans and places the package on a mobile drive unit to be routed to an appropriate drop point.

There is variation in the arm setup among workcells due to real-world constraints of industrial facilities, and the exact information about the incoming items is generally unknown. For this reason, a perception system has been developed, which aims to provide information regarding the packages on the conveyor belt. Even with effective perception in the loop, however, several challenges remain for successful picking, including:

\begin{itemize}[leftmargin=*]
    \item The packages have different types of material (e.g., rigid or non-rigid, smooth or rough, etc.), requiring different picking strategies. For example, packages can be rigid boxes, deformable polybags, or semi-rigid containers in a mail sorting application.
    \item While the perception system can estimate an object's dimensions and material type, the mass and mass distributions of the incoming objects are more difficult to evaluate.
    \item A typical scene of Robin will contain many packages, often in a pile, with many objects only partially observed or some wholly buried in the pile.
    \item The fleet of robots is generally heterogeneous across workcells. There are variations in the workcell design, the operation environments (e.g., the surface where the packages are picked from or placed on), and the manipulator arm models, and different end-of-arm tools (EoAT) may be used due to changes in hardware over the deployment period.
\end{itemize}

A crucial metric in the robotic induction task is the success rate of picking attempts. Ideally, in a pile of objects, the robot should pick them one by one and place them in the target area without dropping any items. Two significant types of failure are possible: either the robot fails to find a suitable pick in the scene (e.g., due to potential collisions or if the objects are somehow out of reach), or the pick attempt is unsuccessful (e.g., the object is dropped after being picked). We call the former as \textit{planning failure} and the latter as \textit{holding failure}. An additional type of failure arises in item singulation when inadvertently more than one item is picked and placed at the same time. We call this type of failure \textit{multi-pick failure}.

To deal with planning failures, in theory, it is possible to check for reachability and collisions for a large finite number of picks in the scene until a viable pick is found. In practice, however, computing collision-free arm trajectories and performing reachability checks are expensive, and the high throughput requirements of the industrial operation limits the number of picks that can be analyzed for each scene. Even if viable picks are found that pass the checks, there is a need to choose the picks that succeed in holding and transferring the package to the mobile robot.

In this context, this work demonstrates how machine learning models trained on historical pick outcomes from a production system can be leveraged to overcome these challenges and improve the performance of such large-scale deployments. Moreover, as the model's ability to predict outcomes improves with more data, the estimated pick qualities become more accurate over time. In particular, the contributions of this demonstration can be summarized as follows:

\begin{itemize}[leftmargin=*]
    \item We demonstrate a large-scale system for predicting pick qualities using machine learning. During our evaluation, this system picked up to 5~million packages daily (i.e., over 200~million packages over the corresponding period).
    \item We describe a ranking strategy for the picks, which uses the learned pick quality prediction system. This strategy has improved the production system's metrics compared to manually engineered, heuristic methods.
    \item We show that retraining the model on more recent data improves performance, indicating that this learning system is already effective with smaller amounts of data but can also improve over time with more recent and increasing datasets.
\end{itemize}

The rest of this demonstration paper is organized as follows: Section~\ref{sec:related-work} reviews the related work and state of the art; Section~\ref{sec:problem} formalizes the problem considered in this work; 
Sections~\ref{sec:pick-success} and~\ref{sec:ranking} explain the methods for pick success prediction and learned pick ranking used by the demonstrated system;
Section~\ref{sec:tests} describes the evaluation performed and discusses the results of the corresponding tests in the production system; finally, Section~\ref{sec:conclusion} discusses the lessons learned and future efforts in this area.
\section{Related Work} \label{sec:related-work}

Having an induction scene with several objects (short-handed as an \textit{induct}), the robot needs to execute one pick to move an object out of the pile and place it in the desired spot at each step. The robot may consider many candidate picks, but at each stage needs to choose and execute just the one with the highest predicted chance of success. 

Ideally, the highest-ranking candidate pick should be the one chosen for execution, eliminating the need for suggesting more than one candidate pick. However, a complete feasibility evaluation of all potential candidates is usually impractical due to computational and time constraints. Therefore, an ordered list of candidate picks should be provided to the robot; the robot will evaluate the candidate picks one by one and execute the highest-ranked candidate that passes the feasibility checks (e.g., collision and robot arm reachability checks). 

Various strategies can be devised to order the candidate picks. Some of these methods are based on hand-crafted intuitive heuristics, such as prioritizing the larger objects or the objects at the top of the pile and preferring the picks closer to the center of the objects and the ones having a higher number of activated suction cups during EoAT's contact with the object. In practice, such heuristics may work for a nominal induct but fail in complex scenarios and edge cases. Trying to manually handle all possible scenarios with more heuristics quickly becomes intractable. As an alternative, we can consider a data-driven approach that uses machine learning to \emph{learn} a metric for the success probability of a particular pick and then rank the picks based on this score. 

Pick selection based on a score learned from data has been an active research area in the past two decades. \citet{gg-cnn} learn to estimate grasp qualities, angles, and gripper widths for each pixel in a depth image, assuming that the parallel jaw gripper's center is aligned with that pixel. The learning model is trained using a dataset of actual and simulated grasps. However, it does not work for non-vertical grasps and requires only a single object in its region of interest. 

\citet{geometric} use a set of visual and geometric features with K-Nearest Neighbor clustering to predict grasp success for BarrettHand\texttrademark. However, it is only suitable for 2-D planar shapes and does not generalize to the 3-D cluttered scene of industrial inducts. 

\citet{single-shot} propose a learning method for simultaneous object detection, semantic segmentation, and grasping detection. However, as a black box, it is difficult to improve the grasp quality using this method. A more modular approach would allow simpler debugging for settings outside the lab's controlled environment.

\citet{mahler2016dex, dex-net} show the feasibility of directly predicting grasp qualities from point clouds with sufficient training data. It is designed for a gripper and a single suction cup at the EoAT. Similar to the work by \citet{single-shot}, the biggest drawback of this method is that the output is hard to interpret, and it is difficult to intelligently improve the performance beyond increasing the training dataset size. However, many ideas from this work inspired our solution.

\citet{multi-affordance} focus on robotic manipulators with multiple EoATs. They propose having various sets of grasps for different EoAT types based on the robot's perceived environment and selecting the EoAT based on the highest predicted success probability. It considers each affordance (e.g., which suction cups are active or inactive on the EoAT) as context rather than as a feature, producing many grasp sets to evaluate against the scene when there are many affordances. This drawback makes the approach infeasible for real-time applications with EoATs that have many affordances.

Some scenes may be void of any picks estimated to succeed with high probability. In this case, \citet{affordance} propose a novel interactive exploration strategy that learns to push the objects around to obtain a better set of possible grasps in a complicated environment. While this method can potentially help in scenarios with no feasible picks or grasps, it is time-consuming and cannot be directly deployed in fast-paced industrial tasks.

Most of the research on grasp and pick quality and success prediction has been done in controlled lab settings, allowing to hold assumptions such as having singulated, rigid, or 2-D shaped objects or only performing vertical picks. In order to have a working system in real-world uncontrolled settings, such as a package fulfillment center or a mail package sortation facility, the method should be able to overcome these limiting assumptions. 

In our proposed method, we have identified a selection of relevant features and have developed models that can assess the pick's quality in a cluttered uncontrolled scene without the limitations of other methods and within the industrial computational and timing constraints. The methods are deployed across a fleet of Robin manipulator robots in fulfillment centers and have been responsible for picking over 200~million packages during our evaluation period. To the best of our knowledge, our work is the only method for predicting the pick quality and ranking of picks that can work with different EoAT orientations, uncertain object material and properties, and cluttered environments.
\section{Problem Statement} \label{sec:problem}

Consider the picking task illustrated in Figure~\ref{fig:robin-workcell}. The task is initiated when a scene of cluttered packages of different types arrives at a reachable area via a conveyor belt. The conveyor belt and the scene remain static throughout the picking process until the scene is ``cleared,'' which occurs when no reachable packages remain or when some exception occurs, and the next scene arrives. 

The action of picking is performed by an induction manipulation robot consisting of a multiple-DoF arm with an end-of-arm tool (EoAT). The EoAT may consist of one or more suction cups. Depending on the EoAT design, each suction cup may be controlled individually, only as groups, or only all together.

Each \textit{pick} is defined as a set of variables determining the actions of the robot: a 3-D point in space (i.e., the desired pick point where the EoAT makes contact with an item's surface), the desired 3-D orientation of the EoAT at the pick point, and a set of desired active suction cups on the EoAT. Note that a single package or even a package segment may be associated with many candidate picks.

At each time $t$, we will use $x_t$ to represent the state of the current scene and $\mathcal{P}_t$ to denote the complete set of possible picks over the scene, determined by an elementary filtering process (e.g., making sure the pick point is on an item). 

Given a scene $x_t$ and any pick $p \in \mathcal{P}_t$, we can construct a $d$-dimensional feature vector $\phi(x_t, p) \in \mathbb{R}^d$ that encodes not only the parameters defining the pick $p$ but also how the pick relates to the scene. For example, it may include the distance from the bottom of each suction cup to the surfaces of the packages beneath, estimated from point-cloud data. See Section~\ref{sec:pick-success:features} for the extracted features used in our current deployment.

The role of the pick ranker is to use $\phi(x_t, p)$ for each $p \in \mathcal{P}_t$ to define an ordering over the candidate picks $\mathcal{P}_t$. This ordered list is passed through an final filtering step (e.g., checking the feasibility of planning the arm motion without a collision), and the robot executes the first feasible pick. The scene is cleared if no viable picks are found (i.e., \textit{planning failure}). Once a feasible pick is executed, whether successful or not, the process starts all over again on the next scene, which may be a slightly modified or entirely new scene.

Ideally, the pick ranker would have perfect knowledge of the final filtering step and would dictate just a single pick to be executed. In practice, the final filtering process can vary by location and other constraints of the particular deployed robot, which may not be known beforehand. To minimize complexity, we only consider memoryless pick ranking systems: only the current scene is considered when choosing a pick, and there is no effort to plan ahead a sequence of picks. 

We assume there exists a function $F : \mathbb{R}^d \rightarrow [0,1]$ such that for a scene $x_t$ and any pick $p \in \mathcal{P}_t$ the probability that a pick $p$ will be successful is equal to $F( \phi(x_t, p) ) \in [0,1]$.
Note that this model assumes $\phi(x_t, p)$ contains all necessary information about whether a pick will be successful. This simplifying assumption does not reflect that there may be unobserved factors influencing pick success, such as the weight distribution inside the package. Extending our model to handle such partially observed settings is ongoing work. 

To maximize the probability of a successful pick at time $t$, an ideal pick ranker would rank the picks of $\mathcal{P}_t$ in decreasing order of $F( \phi(x_t, p) )$.
Note that under this model, the success probability $F( \phi(x_t, p) )$ is agnostic to picks $p$ that do not pass the final filtering process and thus can take an arbitrary value.

In practice, we evaluate a surrogate for $F( \phi(x_t, p) )$ (which takes a non-negligible amount of computation), so ideally, $\mathcal{P}_t$ would only include those picks that pass the final filtering process.

Of course, the true $F$ is unknown, but we can estimate it with data and an appropriate machine learning model (see Section~\ref{sec:pick-success}).

\section{Learning to Estimate Pick Success} \label{sec:pick-success}

This section describes our data-driven approach to estimating the probability of success for a given pick in a scene. 
First, we provide the details for the training datasets used in our work, and then we briefly outline the features extracted for our model. Finally, we explain the details of the models developed for this project.

\subsection{Training Dataset} \label{sec:pick-success:dataset}

We compiled three datasets from hundreds of actual induction cells in Amazon fulfillment centers. Due to the nominal success of pick ranking heuristic methods used in the past, an independent and identically distributed (IID) random draw of inducts would lead to a severely imbalanced dataset, with pick success examples vastly outnumbering the failure examples. Therefore, we oversampled failures in all the training datasets to create a more balanced dataset.

\begin{itemize}[leftmargin=*]
    \item \textbf{TrainDataset-Center:} This dataset contains \textmidtilde395K robotic inducts composed of 335,226 successful and 59,646 failed examples. Additionally, due to the used heuristics, the location of the picks in the dataset is as close to the center of package segments as possible. 

    \item \textbf{TrainDataset-Random:} For this dataset, \textmidtilde41K randomly selected inducts from \textit{TrainDataset-Center} are replaced with new inducts that were randomly distributed to be picked anywhere on the packages' segments with a higher chance of being close to center when a center pick is possible. The new set of inducts comprises 34,715 successful and 6,673 failed picks, and the total size of the newly-created dataset is the same as \textit{TrainDataset-Center}.
    
    \item \textbf{TrainDataset-Past:} This dataset is compiled from historical data collected from inducts executed before the timeframe of \textit{TrainDataset-Center} and \textit{TrainDataset-Random} inducts. It contains \textmidtilde230K inducts composed of 195,408 successful and 34,482 failed examples. Similar to \textit{TrainDataset-Center}, the location of the picks in the dataset is as close to the center of package segments as possible.
\end{itemize}

Each induction consists of the RGB image data captured by a camera at the top of the workcell looking straight down, depth images, and metadata. The metadata includes information on the induct, such as the ground truth on the success or failure, as well as information about the workcell (e.g., the station code, the type of manipulator arm and EoAT).

\subsection{Feature Extraction} \label{sec:pick-success:features}

\begin{figure}[!htb]
    \centering
    \includegraphics[width=\linewidth]{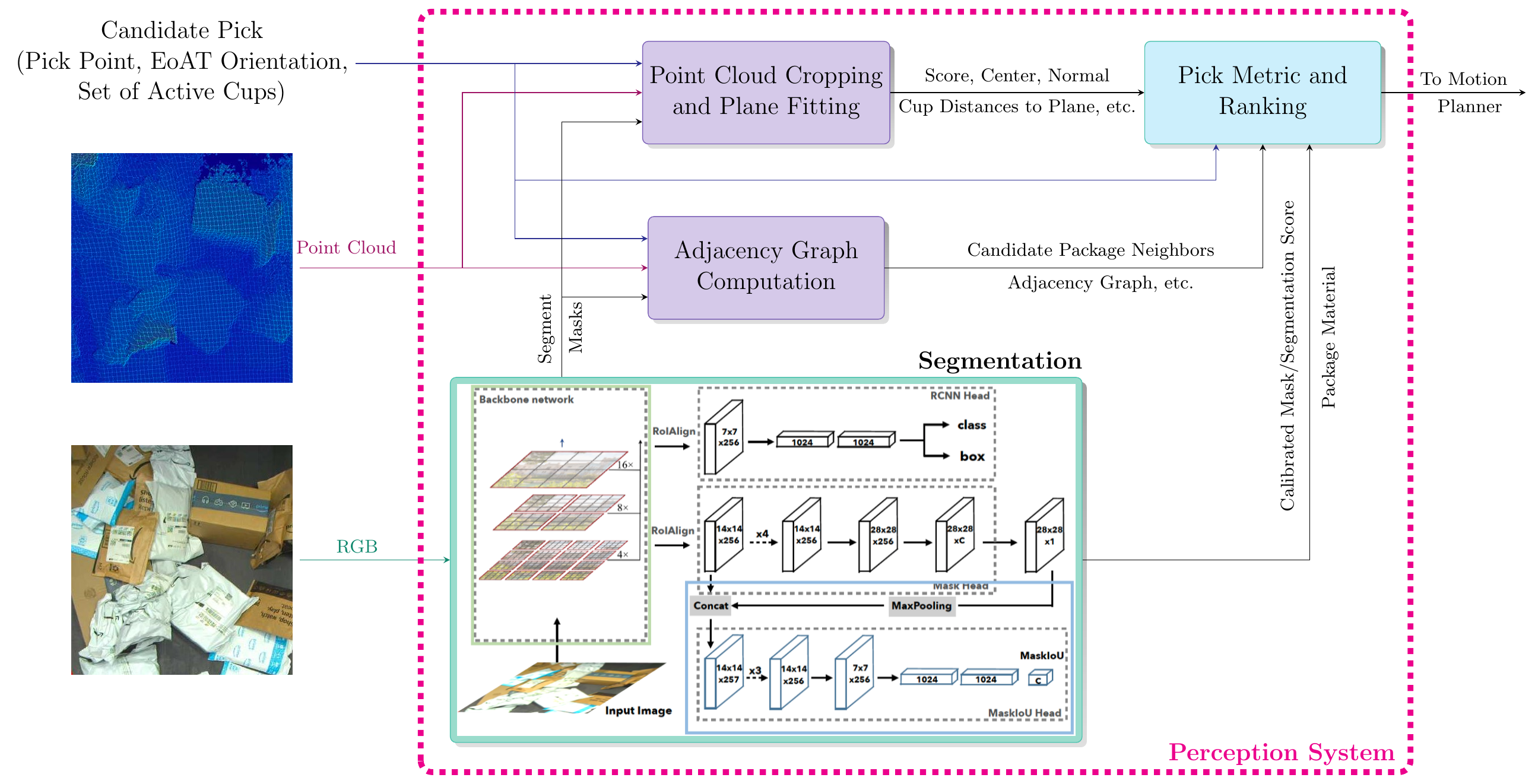}
    \caption{An overview of the perception system design to extract features for the learned pick success model.}
    \label{fig:system-diagram}
\end{figure}

We compute a set of features for each induct using the metadata, RGB, and depth images. Specifically, the camera data is processed by our perception system to generate segments of the packages and tag each segment with an associated package type label. Additional statistics are computed for each segment using depth information (e.g., surface normals and the quality of plane fitting). An overview of our perception system design to extract the required features is shown in Figure~\ref{fig:system-diagram}. 

The segmentation module is a deep network based on Mask Scoring R-CNN~\cite{mask-scoring-rcnn} with a Swin-T backbone~\cite{swin-t} for predicting the package material, instance segmentation masks, and the classification and segmentation scores. The rest of the features extracted for the model are directly computed from the input pick and point-cloud data. 

Based on our feature importance studies, we identified the following features as significant predictors:

\begin{itemize}[leftmargin=*]
    \item \textbf{Package height:} We believe this feature correlates with the package's momentum and, therefore, can impact the shear force at the suction cups and the pick's stability.
    \item \textbf{Quality of plane fitting:} We fit a plane on each segment, and we speculate that a better plane fit correlates with a better seal between the suction cups and the package.
    \item \textbf{Number of activated suction cups:} More active suction cups can mean a more stable pick, reducing the failure probability.
    \item \textbf{Alignment quality between the suction cups and the package surface:} This feature is computed as the offsets between the package surface normal vector and the normal vector of the suction cups. We also expect this feature to be significant since a better alignment indicates a better seal between the suction cups and the package surface.
\end{itemize}

\begin{figure}[!htb]
    \centering
    \includegraphics[width=0.8\linewidth]{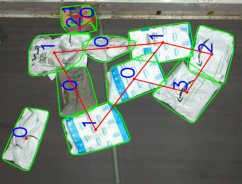}
    \caption{Example of an adjacency graph for a cluster of items.}
    \label{fig:adj-graph}
\end{figure}

In addition to the above and other segment-specific features, we compute features that describe each segment's relationship with its surroundings, including the number of nearby segments and the \textit{adjacency graph} features. To compute the adjacency graph features, we construct a graph that captures the topological order of the package segments. This graph captures each detected segment's relative height with respect to its adjacent neighbor segments. Figure~\ref{fig:adj-graph} shows an example where the numbers represent the relative position ranking of the segment among its neighbors.

\subsection{Pick Success Model} \label{sec:pick-success:model}

The features described in Section~\ref{sec:pick-success:features}, along with the ground truth knowledge of induct success or failure, were extracted for the inducts in the training datasets (Section~\ref{sec:pick-success:dataset}). The AutoGluon library~\cite{autogluon} was leveraged for training, model selection, and hyperparameter tuning for the pick success prediction binary classification task. Many models showed similar performance, but a gradient boosting tree, specifically a CatBoost model~\cite{catboost}, was among the top performers and was chosen for our implementation. We also evaluated other machine learning libraries and modeling options, such as multilayer perceptron (MLP) with Scikit-learn~\cite{scikit-learn}; however, models trained with AutoGluon showed superior performances.

In contrast to our strategy of extracting interpretable (tabular) features, prior works have trained models that directly predict the pick success from some combination of the input RGB image, depth, and pick features. We also benchmarked that approach but did not see a significant improvement over our model's performance. Moreover, our method has a few advantages over the pixels-to-prediction approach:

\begin{enumerate}[leftmargin=*]
    \item \textbf{Interpretability:} We found it challenging to understand what made a particular scene easy or difficult with the pixels-to-prediction approach and why a specific pick failed. In contrast, the tabular features we extracted were easy to interpret and allowed us to characterize different failure modes. This helps to identify weaknesses in the training dataset to refine it later.
    
    \item \textbf{Computation:} A gradient boosting tree is much faster to train and evaluate than a typical image classification model. This means that at deployment, we can evaluate many picks very quickly. 
    
    \item \textbf{Uncertainty quantification:} The CatBoost package natively supports sampling models from a posterior, which can be used to generate ensembles. In anecdotal studies, we found that these ensembles capture uncertainty very well. Because our training data does not cover all possible picks for all scenes, capturing such uncertainty helps provide more conservative predictions.
\end{enumerate}

\section{Pick Ranking} \label{sec:ranking}

Let us assume that the robot has selected several picks in the scene. For example, the system may generate one or more picks per each object segment in the robot's region of interest. The model described in Section~\ref{sec:pick-success} can output estimated probabilities of successfully picking up and holding the desired items for each candidate pick. Assuming the estimate closely correlates with the actual probability of success, the picks with a higher likelihood of success should be prioritized. 

Ranking the picks based on their success probability estimate provides two benefits: the throughput of the workcell is improved due to the higher chance of picking each item up on the first try, and, by picking up the ``easier-to-pick'' items first, the ``harder-to-pick'' items become easier to pick (e.g., the occluding or very close items are removed around them, leaving them singulated). 

In our work, we rank the picks in two steps. First, the picks are grouped by package segments, and the pick success probabilities or other heuristics are leveraged to rank the segments. When picking success probability is employed directly, we estimate the success probability of a segment as $P_{segment} = \max\limits_{i = 1\dots n}{P_{pick}^i}$, where $P_{pick}^i$ is the success probability of the segment's $i^{th}$ pick. Finally, once the segments are ranked, for each segment, we order its picks based on their success probability predictions. The two-step ranking is due to the logistics behind our system structure and the desired flexibility to try different methods for the two steps.

\begin{figure}[!htb]
    \centering
    \includegraphics[width=0.497\linewidth]{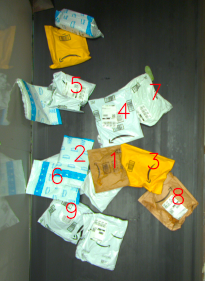}%
    \hfill%
    \includegraphics[width=0.497\linewidth]{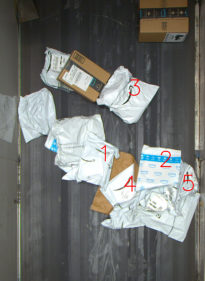}%
    \caption{Examples of ranking the packages based on the highest success probability estimate of their corresponding picks. A smaller rank number represents a higher priority.}
    \label{fig:package-ranking}
\end{figure}

Figure~\ref{fig:package-ranking} shows two examples of package rankings using our model, where the flat and large package segments are prioritized over the crumpled and small ones. 

\begin{figure}[!htb]
    \centering
    \begin{subfigure}{.33\linewidth}
        \includegraphics[width=\linewidth]{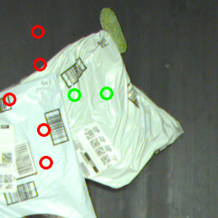}
        \caption{$P_{pick} = 0.708$}
    \end{subfigure}%
    \hfill%
    \begin{subfigure}{.33\linewidth}
        \includegraphics[width=\linewidth]{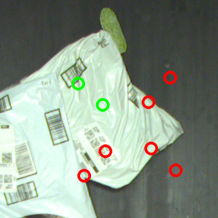}
        \caption{$P_{pick} = 0.756$}
    \end{subfigure}%
    \hfill%
    \begin{subfigure}{.33\linewidth}
        \includegraphics[width=\linewidth]{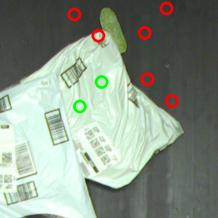}
        \caption{$P_{pick} = 0.825$}
    \end{subfigure}%
    \caption{Estimated success of different picks on a deformable package. The circles correspond to the suction cups. The green and red circles are the active and inactive suction cups.}
    \label{fig:pick-ranking}
\end{figure}

Figure~\ref{fig:pick-ranking} shows the predicted success probability of three picks on a deformable package for picking with the suction cup arrangement illustrated in Figure~\ref{fig:robin:eoat}. All three picks have two activated suction cups on the package. However, the model appears to prioritize the pick configurations where the EoAT is less likely to collide with the surrounding packages and is more likely to succeed.

\section{Experiments and Results} \label{sec:tests}

The methods proposed in Sections~\ref{sec:pick-success} and~\ref{sec:ranking} have been deployed and tested in multiple Amazon sites worldwide. This section presents our testing conditions, experiment results, and our analysis.

\subsection{Hardware} \label{sec:tests:hardware}

\begin{figure}[!htb]
\centering
    \begin{subfigure}[b]{.65\linewidth}
        \includegraphics[width=\linewidth]{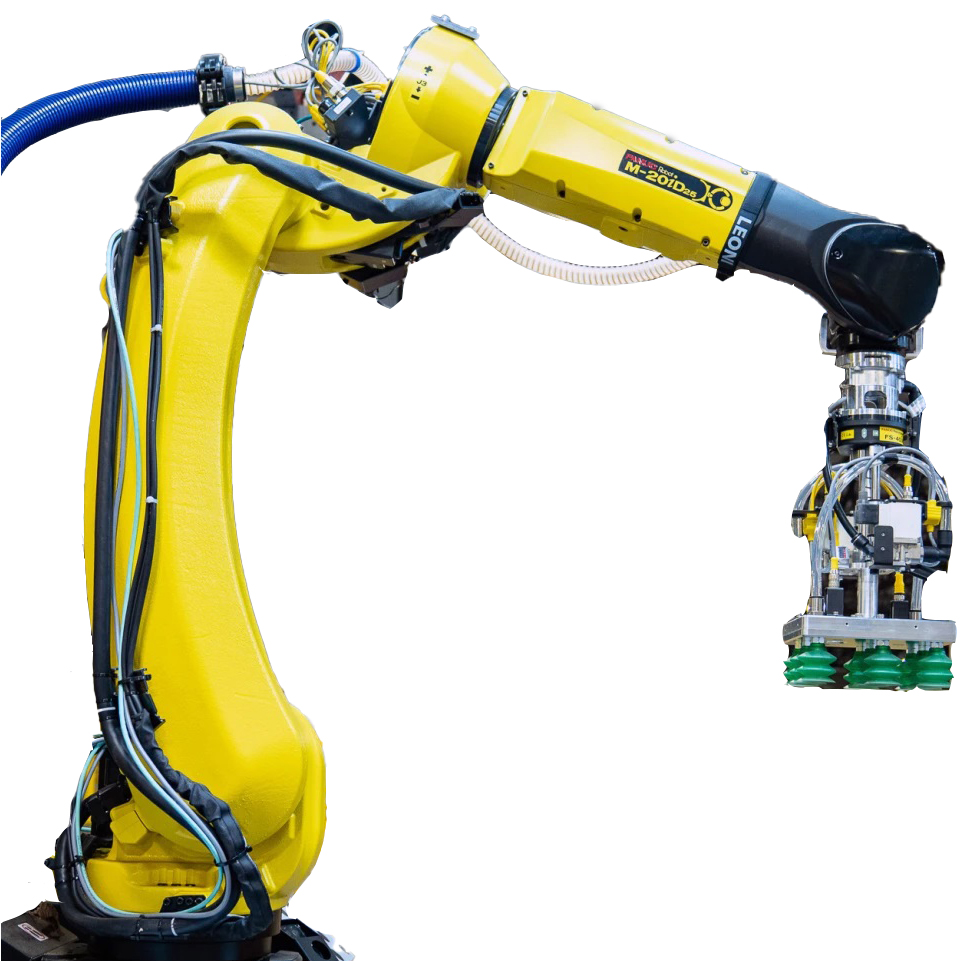}
        \caption{~}
        \label{fig:robin:robot}
    \end{subfigure}%
    \hfill%
    \begin{subfigure}[b]{.34\linewidth}
        \centering
        \includegraphics[width=0.9\linewidth]{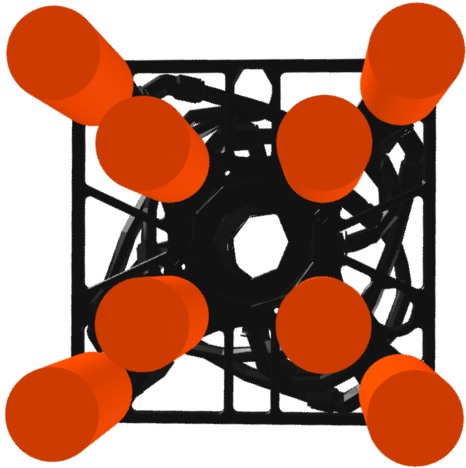}
        \caption{~}
        \label{fig:robin:eoat}
    \end{subfigure}%
    \caption{Robin robotic arm used in our experiments. (a) Manipulator arm. (b) A simulation of the end-of-arm tool design with eight suction cups.}
    \label{fig:robin}
\end{figure}

The proposed method is implemented for Robin robot~\cite{robin} used in Amazon fulfillment centers. The main arm consists of FANUC M-20iD/35 with six controlled axes, 35~kg payload, and 1831~mm reach (Figure~\ref{fig:robin:robot}). The EoAT consists of 8 suction cups arranged in an ``X" configuration with a size of $25\times25$~cm. Each suction cup can be controlled individually. Figure~\ref{fig:robin:eoat} illustrates this EoAT configuration.

\subsection{Baselines and Experiments} \label{sec:tests:baselines}

To evaluate the performance of different \textit{pick success} modeling options, we first consider the following two baselines:

\begin{itemize}[leftmargin=*]
    \item \textbf{AlwaysSuccess:} Always predicting pick success;
    \item \textbf{BoostedTree-Past:} Our pick success model described in Section~\ref{sec:pick-success:model} trained with the historical \textit{TrainDataset-Past} data (see Section~\ref{sec:pick-success:dataset});
\end{itemize}

Historically, our robots were programmed to pick up packages at poses close to the package centers. However, there are cases where the robots must choose picks further away from the package center to avoid collisions, such as with other packages or fixtures on the conveyor belt. To be able to see the effect of choosing off-center picks, we trained two pick success models:

\begin{itemize}[leftmargin=*]
    \item \textbf{BoostedTree-Center:} Our pick success model described in Section~\ref{sec:pick-success:model} trained with the \textit{TrainDataset-Center} data (see Section~\ref{sec:pick-success:dataset});

    \item \textbf{BoostedTree-Random:} Our pick success model described in Section~\ref{sec:pick-success:model} trained with the \textit{TrainDataset-Random} data (see Section~\ref{sec:pick-success:dataset}).
\end{itemize}

\begin{figure}[!htb]
    \centering
    \includegraphics[width=\linewidth]{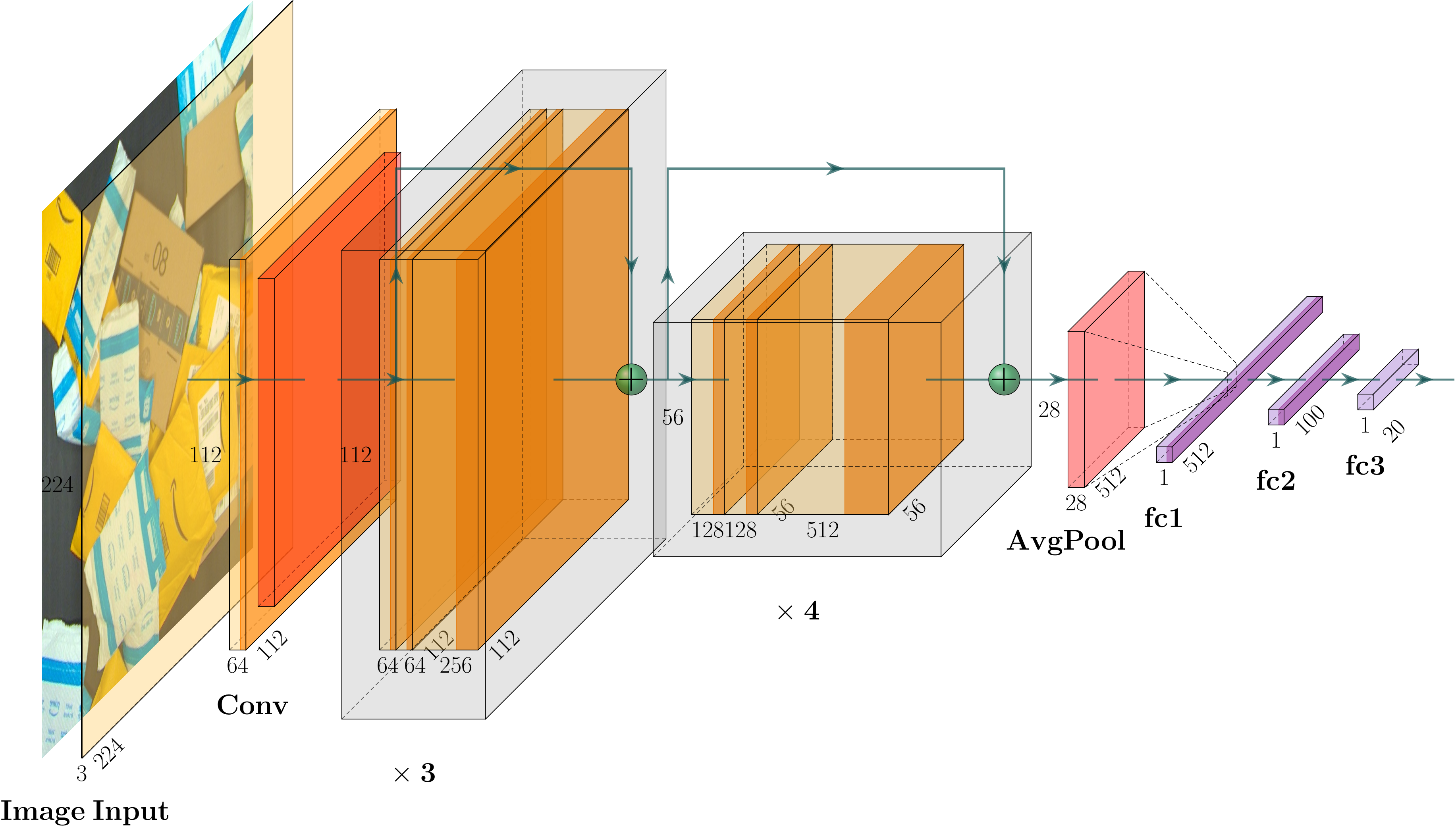}
    \caption{Network architecture for extracting the RGB image embeddings. The input is a $400\times400$ patch around the target pick point (resized to $224\times224$), and the output is the pick success prediction. The blocks before the AvgPool layer are the first three convolution layers of a pre-trained Resnet50 model. The output of the second fully-connected layer (fc2) is used as the embedding. The size of the fc2 layer output (fc3 layer input) is set to 20, 48, or 96 for the desired image embedding sizes.}
    \label{fig:image-network}
\end{figure}

\textbf{CNN-Center:} Finally, we also report the performance of an image-based model, which is a network trained on \textit{TrainDataset-Center} using RGB image crops around the target packages (Figure~\ref{fig:image-network}). 

The following testing datasets are designed to evaluate the pick success estimation models:

\begin{itemize}[leftmargin=*]
    \item \textbf{EvalDataset-Center:} Consisting of \textmidtilde60K picks that are close to the package center. This dataset has an overall pick success rate of 94.40\%.

    \item \textbf{EvalDataset-Random:} Consisting of \textmidtilde38K picks randomly chosen to be anywhere within the package segment. This dataset has an overall pick success rate of 94.01\%.
\end{itemize}

\medskip

For \textit{ranking} segments on the actual robots, we considered several heuristic approaches, including:

\begin{itemize}[leftmargin=*]
    \item \textit{Z-order:} Ranking the picks by the package's target surface elevation. This heuristic is motivated by the assumption that a package whose surface is at the top of the pile is easily reachable, and the robot can avoid collisions or mistakenly pick other occluding packages when trying to pick it up. This heuristic is simple to implement but omits information about actual occlusions and, in practice, fails for unreachable picks and in instances where a portion of the package is at the top of the pile, but the rest is buried under other packages. In addition, since this method is only concerned with the elevation of the package's surface, it is useless for packages lying on the conveyor belt. 
    
    \item \textit{Package size:} Ranking the picks by the package size (or segmentation area). This heuristic assumes that picking packages with a larger visible area will have higher success and that moving out these larger packages first will help declutter the scene, making picking smaller packages easier. A major issue with this heuristic is when smaller packages lie on or overlap with larger ones. This can result in collisions with other packages, difficulty lifting a larger package from under smaller packages, or picking more than one package simultaneously, resulting in holding or multi-pick failures.
    
    \item \textit{Topological order:} Ranking the picks based on the order of occlusion of the packages. Picks to package faces that appear unoccluded get the best score. Picks on package surfaces that are only occluded by unoccluded packages get the next highest score, and so on. This heuristic is moderately simple to implement and prioritizes the unoccluded packages to improve the success probability. However, it fails to recognize the unreachable package surfaces and the occluded packages with unoccluded surfaces, and similar to the Z-order method, it does not differentiate between the picks on the same package surface. Moreover, this method heavily relies on an accurate perception system to compute the segment overlaps.
\end{itemize}

Any combination of the above heuristics may also be employed. For example, a topological order method can be used as the primary ranking criteria, with the Z-order method as the tie-breaker when two packages have the same topological ranks. Some other heuristic approaches are also worth mentioning, such as the measure of how quadrilateral a package is or the confidence score given by the instance segmentation method. These approaches try to indirectly measure a package's occlusion or deformation level but have their own drawbacks and challenges. 

The heuristics above only rank the segments to be picked and cannot differentiate between various picks inside a segment. Once the segments are ranked, a reasonable heuristic approach for ranking picks inside a segment can be giving a higher score to the picks closer to the center of the segment and the ones with a higher number of activated EoAT suction cups.

Having experimented with these heuristics, we establish the following baseline and experiment to evaluate our proposed pick ranking approach:

\begin{itemize}[leftmargin=*]
    \item \textbf{Baseline:} Topological order with Z-order as the tie-breaker for ranking segments; picks with EoAT poses with a higher number of active suction cups, and then the ones with the center of active cups closer to the center of the segment are given higher rank within segments;
    
    \item \textbf{Experiment:} Topological order with learned pick success estimation as the tie-breaker for ranking segments; picks within segments are also ranked with the learned pick success estimation.
\end{itemize}

To evaluate our approach without bias to a particular planning strategy, we allow the robots to select EoAT poses randomly anywhere as long as the poses are within 30~cm distance from the center of the packages on the estimated package surface plane. 

Moreover, additional large-scale experiments were performed through A/B tests on the Robin fleet to assess the performance of different methods. Based on the possible heuristic approaches, we established the following baselines to evaluate against our proposed pick ranking approach:

\begin{enumerate}[leftmargin=*]
    \item \textbf{TopoZ-Center:} Topological order with Z-order as the tie-breaker for ranking segments; picks within segments chosen close to the center of the package's segment with higher rank given to the EoAT poses with a higher number of active suction cups;
    \item \textbf{Z-Center:} Same as \textit{TopoZ-Center}, but only Z-order used directly for ranking segments;
    \item \textbf{TopoZ-Random:} Same as \textit{TopoZ-Center}, but picks within segments are chosen randomly anywhere in the package's segment, with a higher chance for picks closer to the segment's center.
\end{enumerate}

For evaluation of the ranking methods through learned pick success estimation, we designed the following experiments:

\begin{enumerate}[leftmargin=*]
    \item \textbf{TopoLPR-Center:} Topological order with learned pick success estimation as the tie-breaker for ranking segments; picks within segments chosen close to the center of package's segment with learned pick success estimation used for ranking;
    \item \textbf{LPR-Center:} Same as \textit{TopoLPR-Center}, but learned pick success estimation is directly used for ranking the segments; 
    \item \textbf{LPR-Random:} Same as \textit{LPR-Center}, but picks within segments can be chosen randomly anywhere in the package's segment. 
\end{enumerate}

\subsection{Evaluation Metrics} \label{sec:tests:metrics}

Due to the rarity of failed inducts, our evaluation datasets are severely imbalanced. As a result, evaluating the classification \textit{accuracy} does not serve as an informative metric for this task (for example, a baseline that always predicts pick success will get over 94\% accuracy).

On the other hand, the main objective of a pick success classifier is to use its output estimates for ranking the picks so that picks with a higher chance of success are ranked higher.

We choose Receiver Operating Characteristic (ROC) Area Under the Curve (AUC) score as the metric for evaluating the pick success models. Mathematically, the ROC-AUC score is the same as the probability of a classifier ranking a randomly-chosen positive example higher than a randomly chosen negative example, i.e., $P(score(x^+)>score(x^-))$ (see~\cite{roc-auc} for the proof). Therefore, when all the successful picks in the testing datasets are ranked higher than all the failed picks, the ROC-AUC score would be 1.0, and when all the failed picks are ranked higher than successful picks, the ROC-AUC score would be 0. Therefore, the ROC-AUC score is a good metric for evaluating the pick-ranking ability of different models. 

To evaluate the pick ranking system on the actual hardware, we also compute the percentage of picks when the robots fail to transfer a package from the conveyor belts to the mobile robots. We call this metric as \textit{failure rate}.

\subsection{Results} \label{sec:tests:results}

All models introduced in Section~\ref{sec:tests:baselines} were evaluated on both testing datasets \textit{EvalDataset-Center} and \textit{EvalDataset-Random}. Table~\ref{tbl:roc-auc-scores} presents the ROC-AUC scores of these models with confidence intervals.

\begin{table}[!htb]
    \caption{ROC-AUC scores with confidence intervals of different models for the pick success estimation task.}
    \label{tbl:roc-auc-scores}
    \centering
    \begin{tabular}{lcc}
        \toprule
         Model &  EvalDataset-Center & EvalDataset-Random\\
         \midrule
         AlwaysSuccess & 0.5 (0.5, 0.5) & 0.5 (0.5, 0.5)\\
         BoostedTree-Past & 0.725 (0.717, 0.732) & 0.807 (0.799, 0.815)\\
         \cdashlinelr{1-3}
         BoostedTree-Center & 0.755 (0.748, 0.761) & 0.802 (0.792, 0.810)\\
         BoostedTree-Random & 0.758 (0.752, 0.765) & 0.848 (0.840, 0.855)\\
         CNN-Center & 0.570 (0.560, 0.579)  & 0.703 (0.693, 0.712)\\
         \bottomrule
    \end{tabular}
\end{table}

\begin{figure}[!htb]
\centering
    \begin{subfigure}[b]{0.49\linewidth}
        \includegraphics[width=\textwidth]{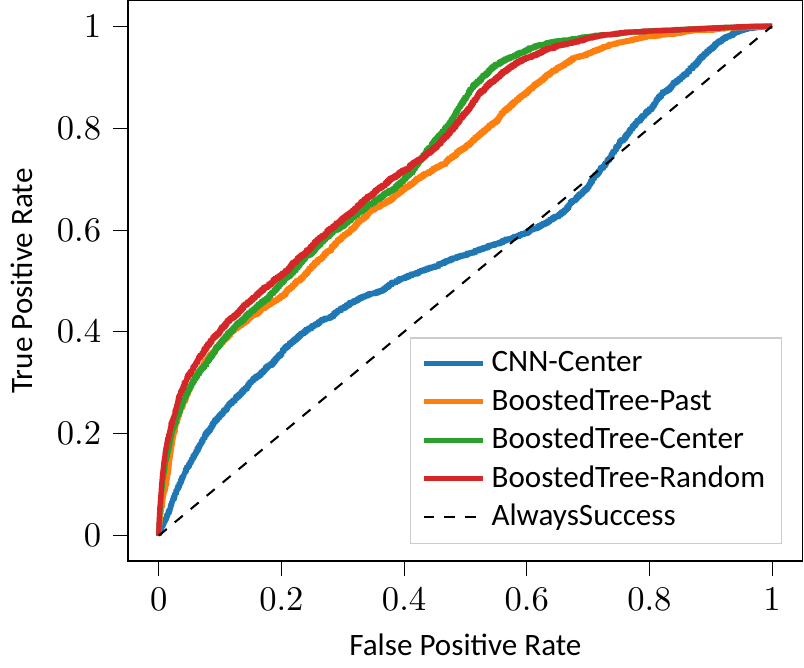}
        \caption{~}
        \label{fig:results-roc-curves:center}
    \end{subfigure}%
    \hfill%
    \begin{subfigure}[b]{0.49\linewidth}
        \includegraphics[width=\textwidth]{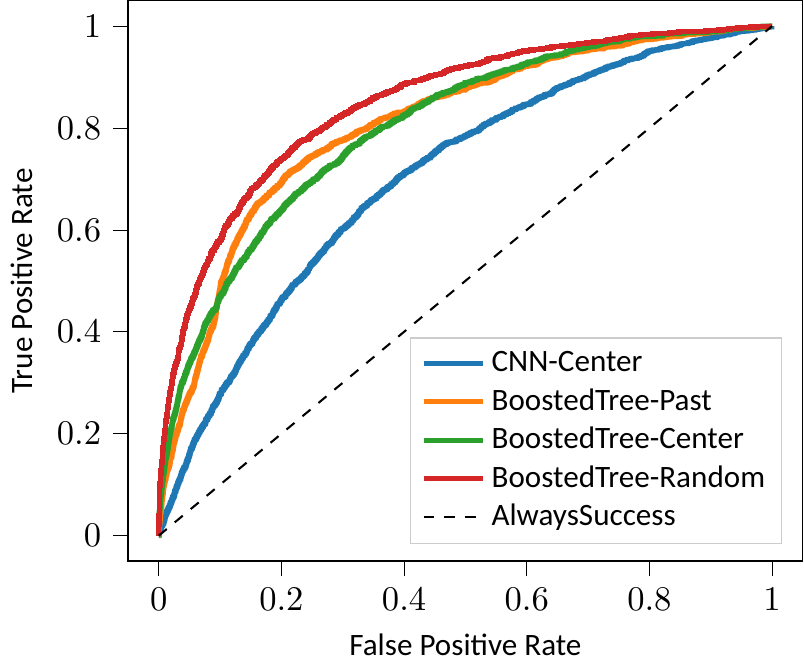}
        \caption{~}
        \label{fig:results-roc-curves:random}
    \end{subfigure}
\caption{ROC curves of all models described in~\ref{tbl:roc-auc-scores} tested on evaluation datasets (a) EvalDataset-Center and (b) EvalDataset-Random.}
\label{fig:results-roc-curves}
\end{figure}

As seen from Table~\ref{tbl:roc-auc-scores}, all the machine learning models beat the naive baseline \textit{AlwaysSuccess} that always predicts pick success. Additionally, all models perform better on \textit{EvalDataset-Random} compared to \textit{EvalDataset-Center}. From Figure~\ref{fig:results-roc-curves}, we observe that the ROC curves on the \textit{EvalDataset-Center} dataset (Figure~\ref{fig:results-roc-curves:center}) are flatter than the ROC curves on the \textit{EvalDataset-Random} dataset (Figure~\ref{fig:results-roc-curves:random}) for middle range false positive rates. This indicates that there are more pick failure examples that are hard to differentiate from the pick success examples in the \textit{EvalDataset-Center} dataset. 

Upon further investigation of the datasets, we found that a sizeable portion of pick failure examples is due to factors not included in our feature set, such as suction cup degradations. We believe the proportion of such examples is more significant when the robot action space is more constrained, such as always attempting to pick packages close to the center.

When comparing models trained with data from different time ranges (i.e., \textit{BoostedTree-Past} vs. \textit{BoostedTree-Center} and \textit{BoostedTree-Random}), we observe the model performance improves when it is trained with more recent data. 

Additionally, we find that the model trained with picks sampled anywhere within the segment (i.e., \textit{BoostedTree-Random}) slightly outperforms the model trained with picks always close to the center when evaluated on \textit{EvalDataset-Center}. On the other hand, there is a more significant margin between \textit{BoostedTree-Random} and \textit{BoostedTree-Center} when they are evaluated on \textit{EvalDataset-Random}. This shows that our pick success model can interpolate between centered picks and picks that are more off-center, and including the off-center picks in the training dataset helps with predicting pick success for larger varieties of picks, which can be beneficial for packages that are hard to reach and the robot has to select from a list of off-center picks. Finally, the RGB image-based model \textit{CNN-Center} performs much worse than the other modeling options. Given \textit{CNN-Center} makes predictions only based on the package appearance, it shows that additional information about the picks is critical even if the robots attempt to pick up the packages close to the center. 

The \textit{Experiment} ranking method (see Section~\ref{sec:tests:baselines}) was deployed for production on Amazon's fulfillment center robotic fleet and has been used for picking up over 200~million inducts with a success rate of 98\%. We analyzed \textmidtilde180K random robotic inducts performed using \textit{Baseline} and \textit{Experiment} ranking methods to validate our proposed method. Table~\ref{tbl:ab-small-results} summarizes the results, which shows that \textit{Experiment} method improves the pick success rate by about 1.18\%. This 23.7\% reduction in failures, when deployed at a large scale (e.g., millions of picks per day performed on our fleet), has a significant impact on the operation costs.

\begin{table}[!htb]
   \caption{Test results for validation of pick-ranking learned method.}
   \label{tbl:ab-small-results}
   \centering
   \begin{tabular}{lcccc}
     \toprule
    Method & Total Picks & Pick Success & Pick Failure & Success Rate \\
     \midrule
     Baseline & 89,162 & 84,718 & 4,444 & 95.02\%\\
     Experiment & 90,127 & 87,700 & 3,427 & 96.20\%\\
     \bottomrule
   \end{tabular}
\end{table}

To better understand the cases where these methods rank the segments differently, we present a qualitative comparison for two cases that assist with understanding the behavior of the learned pick success method. In general, given similar conditions for two packages, the learned pick success estimation method seems to prefer flatter surfaces and packages with less occlusion while disliking the packages close to the conveyor wall or at hard-to-reach angles.

\subsubsection{Case 1} \label{sec:tests:results:case-1}

\begin{figure}[!htb]
\centering
    \begin{subfigure}[b]{0.495\linewidth}
        \includegraphics[width=\textwidth]{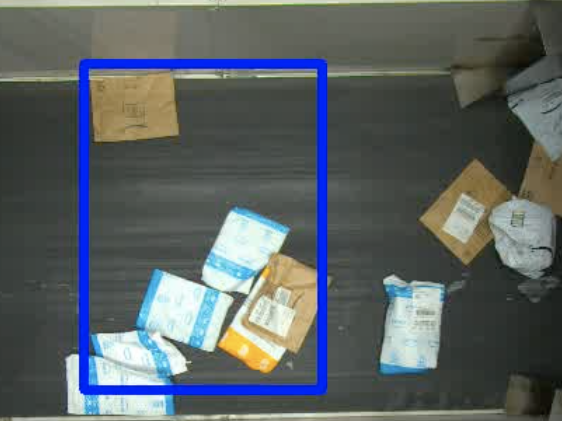}
        \caption{~}
    \end{subfigure}%
    \hfill%
    \begin{subfigure}[b]{0.495\linewidth}
        \includegraphics[width=\textwidth]{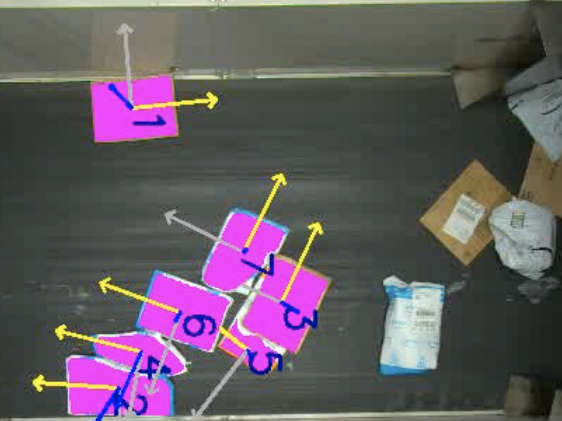}
        \caption{~}
    \end{subfigure}%
    
    \medskip
    \begin{subfigure}[b]{0.495\linewidth}
        \includegraphics[width=\textwidth]{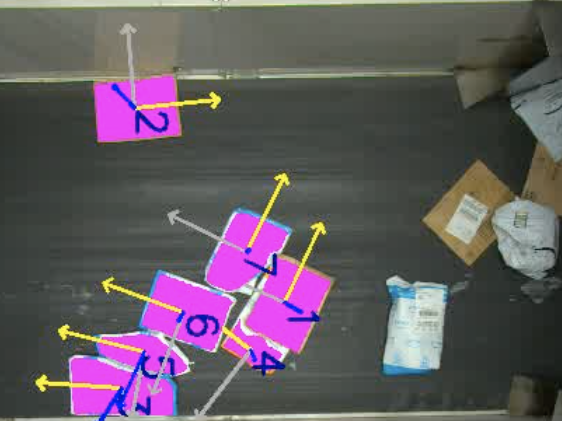}
        \caption{~}
    \end{subfigure}%
    \hfill%
    \begin{subfigure}[b]{0.495\linewidth}
        \includegraphics[width=\textwidth]{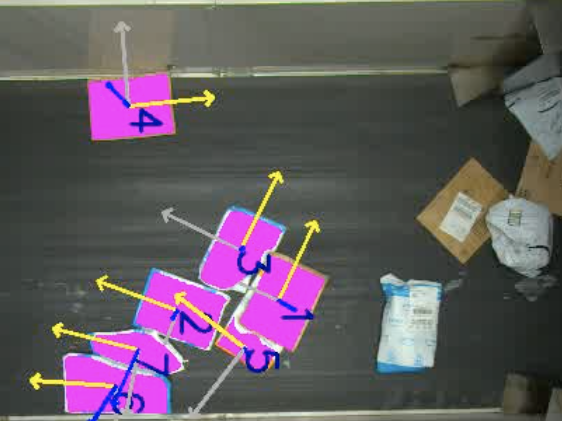}
        \caption{~}
    \end{subfigure}

\caption{Qualitative comparison of different segment ranking methods. (a) An example induct scene with a marked ROI (the blue rectangle). Results for: (b) topological order of segments with Z-order used for tie-breaking, (c) topological order of segments with learned pick success estimation used for tie-breaking, and (d) learned pick success estimation method directly used for segment ranking.}
\label{fig:qualitative-results-1}
\end{figure}

Figure~\ref{fig:qualitative-results-1} shows an induct with its segment ranking results for three methods: topological order with Z-order for tie-breaking, topological order with learned pick success estimation method for tie-breaking, and the learned pick success estimation method. It can be seen that the learned approach prioritizes the packages closer to the center of the conveyor belt (away from the conveyor walls), where the robot is less likely to have difficulties with reaching the pick at the desired angle or colliding with the conveyor wall.

\subsubsection{Case 2} \label{sec:tests:results:case-2}

\begin{figure}[!htb]
\centering
    \begin{subfigure}[b]{0.495\linewidth}
        \includegraphics[width=\textwidth]{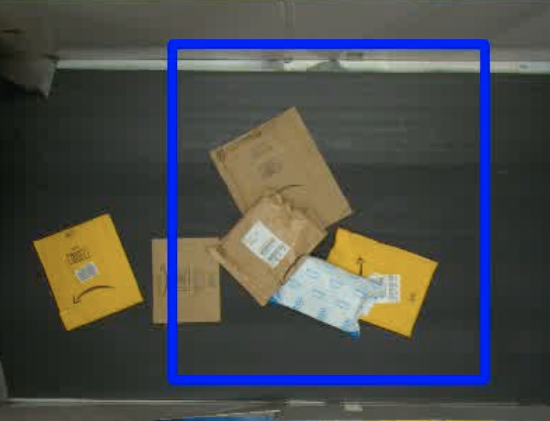}
        \caption{~}
        \label{fig:qualitative-results-2-roi}
    \end{subfigure}%
    \hfill%
    \begin{subfigure}[b]{0.495\linewidth}
        \includegraphics[width=\textwidth]{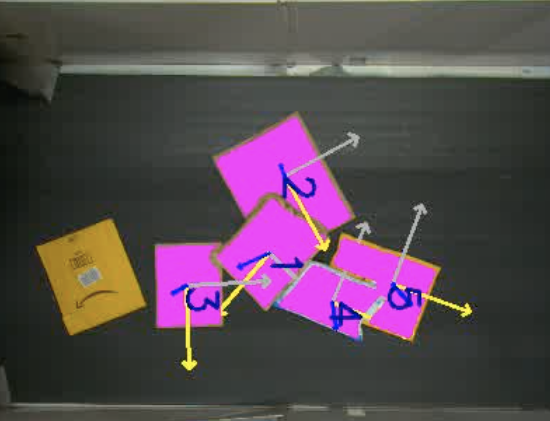}
        \caption{~}
        \label{fig:qualitative-results-2-topo}
    \end{subfigure}%
    
    \medskip
    \begin{subfigure}[b]{0.495\linewidth}
        \includegraphics[width=\textwidth]{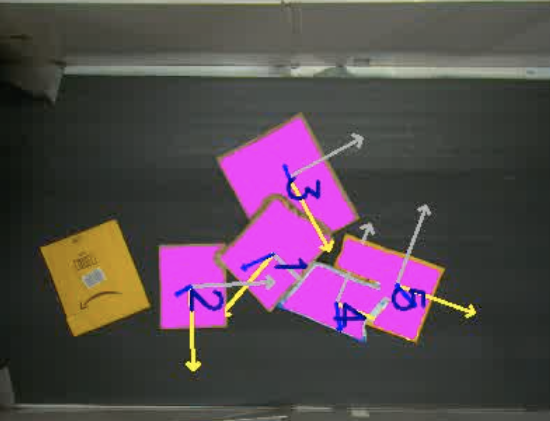}
        \caption{~}
        \label{fig:qualitative-results-2-topolgr}
    \end{subfigure}%
    \hfill%
    \begin{subfigure}[b]{0.495\linewidth}
        \includegraphics[width=\textwidth]{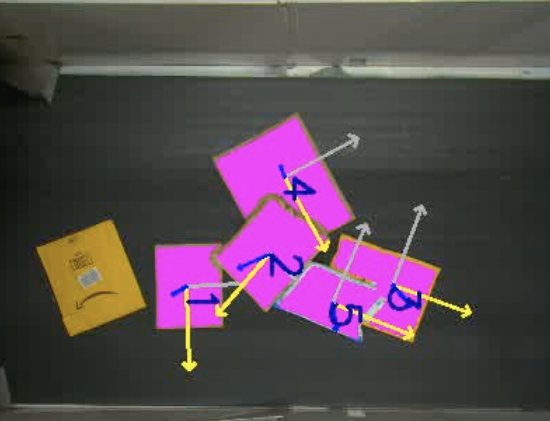}
        \caption{~}
        \label{fig:qualitative-results-2-lgr}
    \end{subfigure}

\caption{Qualitative comparison of different segment ranking methods. (a) Another example induction scene with a marked ROI (the blue rectangle). Results for: (b) topological order of segments with Z-order used for tie-breaking, (c) topological order of segments with learned pick success estimation used for tie-breaking, and (d) learned pick success estimation method directly used for segment ranking.}
\label{fig:qualitative-results-2}
\end{figure}

Figure~\ref{fig:qualitative-results-2} shows an induct with its segment ranking results for the same methods as Case~\hyperref[sec:tests:results:case-1]{1}. Comparing the ranking for segments ranked 2 and 3 in Figures~\ref{fig:qualitative-results-2-topo} and~\ref{fig:qualitative-results-2-topolgr}, it is evident that the learned method prioritizes the packages with less occlusion where the collision with the occluding packages is less likely. On the other hand, when the constraints of topological order are removed, it can be seen in Figure~\ref{fig:qualitative-results-2-lgr} that the learned method slightly prefers the packages with flatter surfaces (the box over the deformable packages) where the chances of holding failure are lower.

Finally, to find the best method, the large-scale A/B test was deployed across the fleet with a small percentage of total inducts allocated to each experiment group in Section~\ref{sec:tests:baselines}. Table~\ref{tbl:ab-large-results} summarizes the results of the A/B experiments.

\begin{table}[!htb]
   \caption{Results of A/B experiments for pick-ranking methods.}
   \label{tbl:ab-large-results}
   \centering
   \begin{tabular}{lccc}
     \toprule
     Method & Total Picks & Failed Picks & Success Rate \\
     \midrule
     TopoZ-Center & 1,158,353 & 89,378 & 92.28\% \\
     Z-Center & 1,157,739 & 89,866 & 92.24\%\\
     TopoZ-Random & 1,158,479 & 109,193 & 90.57\%\\
    \cdashlinelr{1-4}
     TopoLPR-Center & 1,156,697 & 83,535 & 92.78\%\\
     LPR-Center & 1,160,005 & 72,789 & 93.73\%\\
     LPR-Random & 1,157,342 & 79,820 & 93.10\%\\
     \bottomrule
   \end{tabular}
\end{table}

The results show that using the learned pick success estimation to rank the segments (i.e., TopoLPR-Center and LPR-Center) improves the pick success for the robots compared with the manual heuristic ranking methods (i.e., TopoZ-Center and Z-Center). Interestingly, the best improvement comes from the more aggressive approach where we directly apply the learned pick success estimation for ranking (LPR-Center). Given that the heuristic ranking methods heavily depend on the heights of the packages, this suggests that promoting high packages can lead to more challenging picks, such as tall but unstable packages. On the other hand, the pick success model considers the package height only as an input feature along with other information such as the package position, surface normal, and adjacency graph features. Therefore, the pick success model can reason better and deprioritize unstable packages. It is also worth noting that if the picks are chosen randomly, the pick success improvement from the heuristic ranking method (TopoZ-Random) to the learned pick success estimation ranking method (LPR-Random) is even more significant (i.e., from 90.57\% to 93.10\%).

The pick success estimation model deployed in these A/B experiments predicts the pick success probability by taking the average of the predictions from five CatBoost models. The number of trees in the five CatBoost models is 2236, 1069, 799, 1464, and 1208 respectively. For all five models, the depth of the trees is 6, and we used a learning rate of 0.05.

\section{Conclusion} \label{sec:conclusion}

In this paper, we presented a large-scale deployed system for package manipulation, which estimates the pick success using a machine learning model. We demonstrated the effectiveness of this system by evaluating it on over 200~million picks and comparing it to heuristic baselines.

We believe that the recent developments in vision transformers combined with a large amount of induction data from our robotic fleet can improve our image-based network and may provide more valuable image embeddings, enhancing the prediction quality of the overall method. 

Additionally, going through the mistakes made by our model, we realized that a sizable portion of them are due to hardware issues such as a dysfunctioning suction cup. In the future, we intend to leverage the developed pick success model for monitoring the health and analyzing the errors in our robot fleet.

\section*{Acknowledgments}

The work would not have been possible without the support of the wider Amazon Robotics team. More specifically, the authors would like to thank previous and current members of the Robin and Janus teams, who established the underlying technology, provided support, and helped shape and deploy these ideas. The authors would like to give special thanks to David Oreper and Sicong Zhao for enabling deployment and to Shuai Han, Qiujie Cui, Mansour Ahmed, and Andrew Marchese, whose help was critical in the realization of this work.

\section*{Demonstration at RSS 2023}

During the presentation of this paper, we intend to have a live demonstration of our evaluation or production workcells. The demonstration will present live webcams from different sites across the globe showing Robin robots picking and placing incoming packages.

\addtolength{\textheight}{-9.0cm}
\bibliographystyle{plainnat}
\bibliography{references}

\begin{thebibliography}{20}
\providecommand{\natexlab}[1]{#1}
\providecommand{\url}[1]{\texttt{#1}}
\expandafter\ifx\csname urlstyle\endcsname\relax
  \providecommand{\doi}[1]{doi: #1}\else
  \providecommand{\doi}{doi: \begingroup \urlstyle{rm}\Url}\fi

\bibitem[Amazon Science()]{robin}
Amazon Science.
\newblock
  \href{https://www.amazon.science/latest-news/robin-deals-with-a-world-where-things-are-changing-all-around-it}{{Amazon
  Robotics: Robin}}, 2022.
\newblock URL
  \url{https://www.amazon.science/latest-news/robin-deals-with-a-world-where-things-are-changing-all-around-it}.

\bibitem[Araki et~al.(2022)Araki, Hirakawa, Yamashita, and
  Fujiyoshi]{single-shot}
Ryosuke Araki, Tsubasa Hirakawa, Takayoshi Yamashita, and Hironobu Fujiyoshi.
\newblock
  \href{https://www.tandfonline.com/doi/full/10.1080/01691864.2022.2043183}{{MT-DSSD:
  Multi-task} deconvolutional single shot detector for object detection,
  segmentation, and grasping detection}.
\newblock \emph{Advanced Robotics}, 36\penalty0 (8):\penalty0 373--387, 2022.
\newblock \doi{10.1080/01691864.2022.2043183}.
\newblock URL
  \url{https://www.tandfonline.com/doi/full/10.1080/01691864.2022.2043183}.

\bibitem[Björnsson et~al.(2018)Björnsson, Jonsson, and
  Johansen]{pick-and-place-manufacturing}
Andreas Björnsson, Marie Jonsson, and Kerstin Johansen.
\newblock
  \href{https://www.sciencedirect.com/science/article/pii/S0736584517301758}{Automated
  material handling in composite manufacturing using pick-and-place systems –
  a review}.
\newblock \emph{Robotics and Computer-Integrated Manufacturing}, 51:\penalty0
  222--229, 2018.
\newblock ISSN 0736-5845.
\newblock \doi{10.1016/j.rcim.2017.12.003}.
\newblock URL
  \url{https://www.sciencedirect.com/science/article/pii/S0736584517301758}.

\bibitem[Fakoor et~al.(2020)Fakoor, Mueller, Erickson, Chaudhari, and
  Smola]{autogluon}
Rasool Fakoor, Jonas~W Mueller, Nick Erickson, Pratik Chaudhari, and
  Alexander~J Smola.
\newblock
  \href{https://proceedings.neurips.cc/paper/2020/file/62d75fb2e3075506e8837d8f55021ab1-Paper.pdf}{Fast,
  Accurate, and Simple Models for Tabular Data via Augmented Distillation}.
\newblock In \emph{Advances in Neural Information Processing Systems},
  volume~33, pages 8671--8681, 2020.
\newblock URL
  \url{https://proceedings.neurips.cc/paper/2020/file/62d75fb2e3075506e8837d8f55021ab1-Paper.pdf}.

\bibitem[Hanley and McNeil(1982)]{roc-auc}
J~A Hanley and B~J McNeil.
\newblock \href{https://doi.org/10.1148/radiology.143.1.7063747}{The meaning
  and use of the area under a receiver operating characteristic ({ROC}) curve}.
\newblock \emph{Radiology}, 143\penalty0 (1):\penalty0 29--36, 1982.
\newblock \doi{10.1148/radiology.143.1.7063747}.
\newblock URL \url{https://doi.org/10.1148/radiology.143.1.7063747}.

\bibitem[Huang et~al.(2019)Huang, Huang, Gong, Huang, and
  Wang]{mask-scoring-rcnn}
Zhaojin Huang, Lichao Huang, Yongchao Gong, Chang Huang, and Xinggang Wang.
\newblock \href{https://ieeexplore.ieee.org/document/8953609}{{Mask Scoring
  R-CNN}}.
\newblock In \emph{2019 IEEE/CVF Conference on Computer Vision and Pattern
  Recognition (CVPR)}, pages 6402--6411, 2019.
\newblock \doi{10.1109/CVPR.2019.00657}.
\newblock URL \url{https://ieeexplore.ieee.org/document/8953609}.

\bibitem[Liu et~al.(2021{\natexlab{a}})Liu, Deng, Guo, Fang, Sun, and
  Yang]{affordance}
Huaping Liu, Yuhong Deng, Di~Guo, Bin Fang, Fuchun Sun, and Wuqiang Yang.
\newblock \href{https://ieeexplore.ieee.org/document/8970574}{An Interactive
  Perception Method for Warehouse Automation in Smart Cities}.
\newblock \emph{IEEE Transactions on Industrial Informatics}, 17\penalty0
  (2):\penalty0 830--838, 2021{\natexlab{a}}.
\newblock \doi{10.1109/TII.2020.2969680}.
\newblock URL \url{https://ieeexplore.ieee.org/document/8970574}.

\bibitem[Liu et~al.(2021{\natexlab{b}})Liu, Lin, Cao, Hu, Wei, Zhang, Lin, and
  Guo]{swin-t}
Ze~Liu, Yutong Lin, Yue Cao, Han Hu, Yixuan Wei, Zheng Zhang, Stephen Lin, and
  Baining Guo.
\newblock \href{https://ieeexplore.ieee.org/document/9710580}{Swin Transformer:
  Hierarchical Vision Transformer using Shifted Windows}.
\newblock In \emph{2021 IEEE/CVF International Conference on Computer Vision
  (ICCV)}, pages 9992--10002, 2021{\natexlab{b}}.
\newblock \doi{10.1109/ICCV48922.2021.00986}.
\newblock URL \url{https://ieeexplore.ieee.org/document/9710580}.

\bibitem[Mahler et~al.(2016)Mahler, Pokorny, Hou, Roderick, Laskey, Aubry,
  Kohlhoff, Kr{\"o}ger, Kuffner, and Goldberg]{mahler2016dex}
Jeffrey Mahler, Florian~T Pokorny, Brian Hou, Melrose Roderick, Michael Laskey,
  Mathieu Aubry, Kai Kohlhoff, Torsten Kr{\"o}ger, James Kuffner, and Ken
  Goldberg.
\newblock \href{https://ieeexplore.ieee.org/document/7487342}{{Dex-Net 1.0}: A
  cloud-based network of 3d objects for robust grasp planning using a
  multi-armed bandit model with correlated rewards}.
\newblock In \emph{IEEE International Conference on Robotics and Automation
  (ICRA)}, pages 1957--1964, 2016.
\newblock URL \url{https://ieeexplore.ieee.org/document/7487342}.

\bibitem[Mahler et~al.(2019)Mahler, Matl, Satish, Danielczuk, DeRose, McKinley,
  and Goldberg]{dex-net}
Jeffrey Mahler, Matthew Matl, Vishal Satish, Michael Danielczuk, Bill DeRose,
  Stephen McKinley, and Ken Goldberg.
\newblock
  \href{https://www.science.org/doi/abs/10.1126/scirobotics.aau4984}{Learning
  ambidextrous robot grasping policies}.
\newblock \emph{Science Robotics}, 4\penalty0 (26):\penalty0 eaau4984, 2019.
\newblock \doi{10.1126/scirobotics.aau4984}.
\newblock URL
  \url{https://www.science.org/doi/abs/10.1126/scirobotics.aau4984}.

\bibitem[Morales et~al.(2003)Morales, Chinellato, Fagg, and del
  Pobil]{geometric}
A.~Morales, E.~Chinellato, A.H. Fagg, and A.P. del Pobil.
\newblock \href{https://ieeexplore.ieee.org/document/1249685}{Experimental
  prediction of the performance of grasp tasks from visual features}.
\newblock In \emph{Proceedings 2003 IEEE/RSJ International Conference on
  Intelligent Robots and Systems (IROS 2003) (Cat. No.03CH37453)}, volume~4,
  pages 3423--3428 vol.3, 2003.
\newblock \doi{10.1109/IROS.2003.1249685}.
\newblock URL \url{https://ieeexplore.ieee.org/document/1249685}.

\bibitem[Morrison et~al.(2020)Morrison, Corke, and Leitner]{gg-cnn}
Douglas Morrison, Peter Corke, and J\"{u}rgen Leitner.
\newblock
  \href{https://journals.sagepub.com/doi/10.1177/0278364919859066}{Learning
  robust, real-time, reactive robotic grasping}.
\newblock \emph{The International Journal of Robotics Research}, 39\penalty0
  (2-3):\penalty0 183--201, 2020.
\newblock \doi{10.1177/0278364919859066}.
\newblock URL \url{https://journals.sagepub.com/doi/10.1177/0278364919859066}.

\bibitem[Nguyen et~al.(2021)Nguyen, Kuhn, and Franke]{car-wiring}
Huong~Giang Nguyen, Marlene Kuhn, and Jörg Franke.
\newblock
  \href{https://www.sciencedirect.com/science/article/pii/S2212827120314761}{Manufacturing
  automation for automotive wiring harnesses}.
\newblock In \emph{8th CIRP Conference of Assembly Technology and Systems},
  volume~97, pages 379--384, 2021.
\newblock \doi{https://doi.org/10.1016/j.procir.2020.05.254}.
\newblock URL
  \url{https://www.sciencedirect.com/science/article/pii/S2212827120314761}.

\bibitem[Pedregosa et~al.(2011)Pedregosa, Fabian, Varoquaux, Ga\"{e}l,
  Gramfort, Alexandre, Michel, Vincent, Thirion, Bertrand, Grisel, Olivier,
  Blondel, Mathieu, Prettenhofer, Peter, Weiss, Ron, Dubourg, Vincent,
  Vanderplas, Jake, Passos, Alexandre, Cournapeau, David, Brucher, Matthieu,
  Perrot, Matthieu, Duchesnay, and \'{E}douard]{scikit-learn}
Pedregosa, Fabian, Varoquaux, Ga\"{e}l, Gramfort, Alexandre, Michel, Vincent,
  Thirion, Bertrand, Grisel, Olivier, Blondel, Mathieu, Prettenhofer, Peter,
  Weiss, Ron, Dubourg, Vincent, Vanderplas, Jake, Passos, Alexandre,
  Cournapeau, David, Brucher, Matthieu, Perrot, Matthieu, Duchesnay, and
  \'{E}douard.
\newblock
  \href{https://www.jmlr.org/papers/v12/pedregosa11a.html}{Scikit-learn:
  Machine Learning in {P}ython}.
\newblock \emph{Journal of Machine Learning Research}, 12:\penalty0 2825--2830,
  2011.
\newblock \doi{10.5555/1953048.2078195}.
\newblock URL \url{https://www.jmlr.org/papers/v12/pedregosa11a.html}.

\bibitem[Prokhorenkova et~al.(2018)Prokhorenkova, Gusev, Vorobev, Dorogush, and
  Gulin]{catboost}
Liudmila Prokhorenkova, Gleb Gusev, Aleksandr Vorobev, Anna~Veronika Dorogush,
  and Andrey Gulin.
\newblock
  \href{https://papers.nips.cc/paper/2018/hash/14491b756b3a51daac41c24863285549-Abstract.html}{{CatBoost}:
  unbiased boosting with categorical features}.
\newblock In \emph{Advances in Neural Information Processing Systems},
  volume~31, pages 1--11. Curran Associates, Inc., 2018.
\newblock URL
  \url{https://proceedings.neurips.cc/paper/2018/file/14491b756b3a51daac41c24863285549-Paper.pdf}.

\bibitem[Salahuddin and Lee(2022)]{garment-automation}
Mir Salahuddin and Young-A Lee.
\newblock \emph{\href{https://doi.org/10.1007/978-3-030-91135-5_5}{Automation
  with Robotics in Garment Manufacturing}}, pages 75--94.
\newblock Springer International Publishing, Cham, 2022.
\newblock ISBN 978-3-030-91135-5.
\newblock \doi{10.1007/978-3-030-91135-5_5}.
\newblock URL \url{https://doi.org/10.1007/978-3-030-91135-5_5}.

\bibitem[Sobhan and Shaikat(2021)]{pick-and-place-automation}
Nazib Sobhan and Abu~Salman Shaikat.
\newblock
  \href{https://ieeexplore.ieee.org/abstract/document/9526304}{Implementation
  of Pick \& Place Robotic Arm for Warehouse Products Management}.
\newblock In \emph{2021 IEEE 7th International Conference on Smart
  Instrumentation, Measurement and Applications (ICSIMA)}, pages 156--161,
  2021.
\newblock \doi{10.1109/ICSIMA50015.2021.9526304}.
\newblock URL \url{https://ieeexplore.ieee.org/abstract/document/9526304}.

\bibitem[Zeng et~al.(2022)Zeng, Song, Yu, Donlon, Hogan, Bauza, Ma, Taylor,
  Liu, Romo, Fazeli, Alet, Dafle, Holladay, Morona, Nair, Green, Taylor, Liu,
  Funkhouser, and Rodriguez]{multi-affordance}
Andy Zeng, Shuran Song, Kuan-Ting Yu, Elliott Donlon, Francois~R. Hogan, Maria
  Bauza, Daolin Ma, Orion Taylor, Melody Liu, Eudald Romo, Nima Fazeli, Ferran
  Alet, Nikhil~Chavan Dafle, Rachel Holladay, Isabella Morona, Prem~Qu Nair,
  Druck Green, Ian Taylor, Weber Liu, Thomas Funkhouser, and Alberto Rodriguez.
\newblock
  \href{https://journals.sagepub.com/doi/10.1177/0278364919868017}{Robotic
  pick-and-place of novel objects in clutter with multi-affordance grasping and
  cross-domain image matching}.
\newblock \emph{The International Journal of Robotics Research}, 41\penalty0
  (7):\penalty0 690--705, 2022.
\newblock \doi{10.1177/0278364919868017}.
\newblock URL \url{https://journals.sagepub.com/doi/10.1177/0278364919868017}.

\bibitem[Zhang and Zhang(2021)]{apple-crating}
Kun Zhang and Hua Zhang.
\newblock \href{https://ieeexplore.ieee.org/abstract/document/9389974}{Design
  and implementation of automatic apple crating robot technology}.
\newblock In \emph{2021 IEEE 2nd International Conference on Big Data,
  Artificial Intelligence and Internet of Things Engineering (ICBAIE)}, pages
  617--621, 2021.
\newblock \doi{10.1109/ICBAIE52039.2021.9389974}.
\newblock URL \url{https://ieeexplore.ieee.org/abstract/document/9389974}.

\bibitem[Zhang et~al.(2018)Zhang, Pothula, and Lu]{apple-binning-review}
Zhao Zhang, Anand~Kumar Pothula, and Renfu Lu.
\newblock \href{https://elibrary.asabe.org/abstract.asp?aid=49538&t=3}{A Review
  of Bin Filling Technologies for Apple Harvest and Postharvest Handling}.
\newblock \emph{Applied Engineering in Agriculture}, 34\penalty0 (4):\penalty0
  687--703, 2018.
\newblock ISSN 0883-8542.
\newblock URL \url{https://elibrary.asabe.org/abstract.asp?aid=49538&t=3}.

\end{thebibliography}
\newpage


\end{document}